\documentclass{article}
\usepackage{spconf}

\usepackage{amsfonts}
\usepackage{amsmath}
\usepackage{amssymb}
\usepackage{graphicx}

\makeatletter
\def\bstctlcite{\@ifnextchar[{\@bstctlcite}{\@bstctlcite[@auxout]}}
\def\@bstctlcite[#1]#2{\@bsphack
  \@for\@citeb:=#2\do{%
    \edef\@citeb{\expandafter\@firstofone\@citeb}%
    \if@filesw\immediate\write\csname #1\endcsname{\string\citation{\@citeb}}\fi}%
  \@esphack}
\makeatother 

\usepackage{enumerate}
\usepackage{enumitem}
\setlist{nosep, leftmargin=14pt}

\usepackage{subcaption}
\usepackage{xcolor}
\usepackage{flushend}

\usepackage{booktabs}
\usepackage{siunitx}
\usepackage{siunitx}
\title{Explaining Digital Pathology Models via Clustering Activations}

\definecolor{cluster1}{HTML}{1F77B4}
\definecolor{cluster2}{HTML}{FF7F0E}
\definecolor{cluster3}{HTML}{2CA02C}
\definecolor{cluster4}{HTML}{D62728}
\definecolor{cluster5}{HTML}{9467BD}
\definecolor{cluster6}{HTML}{8C564B}

\newcommand{\myref}[2]{\hyperref[#2]{#1~\ref*{#2}}}

\name{Adam Bajger\textsuperscript{1} \hss Jan Obdržálek\textsuperscript{1} \hss Vojtěch Kůr\textsuperscript{1} \hss
 Rudolf Nenutil\textsuperscript{3} \hss Petr Holub\textsuperscript{2} \hss  Vít Musil\textsuperscript{1} \hss Tomáš Brázdil\textsuperscript{1}}

\address{\textsuperscript{1} Faculty of Informatics, Masaryk University, Brno, Czech Republic \\
\textsuperscript{2} Institute of Computer Science, Masaryk University, Brno, Czech Republic \\
\textsuperscript{3} Masaryk Memorial Cancer Institute, Brno, Czech Republic}

\begin{document}
\bstctlcite{IEEEexample:BSTcontrol}
\maketitle
\begin{abstract}
We present a clustering-based explainability technique for digital pathology models based on convolutional neural networks.
Unlike commonly used methods based on saliency maps, such as occlusion, GradCAM, or relevance propagation, which highlight regions that contribute the most to the prediction for a single slide, our method shows the global behaviour of the model under consideration, while also providing more fine-grained information.
The result clusters can be visualised not only to understand the model, but also to increase confidence in its operation, leading to faster adoption in clinical practice.
We also evaluate the performance of our technique on an existing model for detecting prostate cancer, demonstrating its usefulness.
\end{abstract}

\section{Introduction}
The prevalence of oncological diseases is increasing all around the world, and early detection of cancer is crucial for improving treatment outcomes and survival rates.
Digital pathology, the contemporary approach to cancer detection, involves analysing digitized copies of tissue samples, called Whole Slide Images (WSIs).
WSIs are then displayed in a dedicated computer browser, enabling quick navigation and introspection of the digitized tissue. 
Detection of, e.g, prostate cancer in these digital samples relies on identifying a set of relatively well-established morphological patterns that visually differ from the surrounding healthy tissue \cite{gleason-patterns}.

While the above-mentioned pattern identification is traditionally performed by human experts, i.e., trained pathologists, there has been a surge of research into developing deep learning-based AI models to perform this task~\cite{campanella_clinical-grade_2019, litjens_survey_2017}.
However, the adoption of such models in clinical practice lags significantly behind state-of-the-art algorithms~\cite{van_der_velden_explainable_2022,hagele_resolving_2020}.
An important contributing factor is the reluctance of pathologists to implicitly trust these models, which to them are functionally just opaque black boxes producing a yes/no answer~\cite{bhati_survey_2024}.  
However, their trust and confidence in the model can be significantly increased if we can demonstrate what the model considers when making its decision~\cite{plass_explainability_2023, samek_explainable_2019}.
Such research is the realm of explainable AI.

For example, in~\cite{gallo} the authors showed how to develop a model that not only correctly identifies prostate cancer but also demonstrably does so by utilizing the very patterns used for this purpose by trained pathologists.
To achieve this goal, they relied on an explainability technique known as occlusion, which works by replacing the most salient regions of the input image and observing the effect on predictions produced by the network.
Several techniques (other than occlusion) have been developed to visualize the most salient regions in the image, e.g., CAM-based techniques such as GradCAM~\cite{gradcam} (and its variants) and HiResCAM~\cite{HiResCAM}, or LRP~\cite{LRP}.
All these methods visually highlight the regions in the slide that contribute most to the model's decision to classify the slide as containing cancer tissue.

However, such one-dimensional information is inherently limited in showing us what drives the model's decision process, as it can only highlight the importance of regions throughout the input, without providing any reasoning for the importance being assigned. This differs from what a human would do in this task.  
A trained pathologist examining a scan takes into account various types of tissue and morphological features, painting a much more complex picture.
A method that shows global CNN model behaviour, highlighting discrepancies in the internal representations, would be much more useful in this respect.

\textbf{Our contribution.}
We present a clustering-based technique to visually identify parts of the input image that ``look similar'' to the model.
The method works by clustering feature vectors from convolutional layers of a CNN model, facilitating semantic segmentation of the input and providing interpretable insights into the model's operation.

Note that this segmentation is produced from a model that was never trained to do segmentation, but only to classify the tissue scans according to the presence of cancerous tissue.
While similar approaches have been studied in the past, only very few of these were in the field of digital pathology, which brings its own set of challenges, such as working on WSIs whose dimensions are in the order of hundreds of thousands of pixels per slide and which contain lots of empty space, with just a few regions significant for making a diagnosis.
Our approach is designed to easily handle such slides.
On the example of the prostate model of~\cite{gallo}, we demonstrate how the segmented tissue corresponds to the morphological features of the input.
Finally, we compare our results to saliency map approaches, such as GradCAM, showing that the produced clusters indeed refine the information contained in these maps.
The deep insight into the inner workings of a CNN model not only serves as a robust tool for researchers in the area but also promotes trust in such models, improving the adoption rate of CNN-based models in clinical practice.

\textbf{Related work} Clustering-based approaches have been studied previously, in the context of concept identification. Our approach loosely follows~\cite{ferrari_deep_2018}, where clustering (employing non-negative matrix factorization) is used for concept identification in general images.  In the context of histopathology, a similar approach was taken by~\cite{segmentationfactorization}, where clustering to a high number of clusters was used for weakly supervised tissue segmentation: while the model was not trained to segment the tissue, each of the computed clusters was assigned to one of the tissue types based on manually produced slide annotation.

\vspace{0em minus 0.5em}
\section{Materials and Methods}

Our method is intended for digital pathology models based on convolutional neural networks (CNNs).
These networks take as input a tile from a histopathology slide and produce a prediction, which is typically one of predefined classes (e.g., healthy/cancerous) or a numerical value.

\vspace{0em minus 0.5em}
\subsection{Cluster computation}

Let us fix a CNN model, a convolutional layer $\ell$, and let $w\times h$ be the dimensions and $C$ the number of channels in this layer, respectively.
Now take an input image tile $I$ and let $f^c_{i,j}(I)$ be the output of the model for this tile in channel $c$ of the layer $\ell$ at the spatial location $(i, j)$.
The \emph{feature vector} $v_{i,j}(I)$ at the location $(i,j)$ is then a vector of values for all individual channels
$v_{i,j}(I) = [f^1_{i, j}(I), \ldots, f^C_{i, j}(I)]$.

To cluster a set of tiles, we use non-negative matrix factorization (NMF)~\cite{ferrari_deep_2018}.
Since this method requires the feature vector to be non-negative, we use the outputs of the layer $\ell$ after ReLU.
NMF is parametrized by the number of classes $K$.
We fix a training matrix of feature vectors $V$ of dimensions $n \times C$, i.e., the $m$-th row of $V$ is a feature vector of some tile~$I$ at spatial location $(i,j)$.

On training time, NMF finds non-negative matrices $W$ and $H$ of dimensions $n \times K$ and $K \times C$, respectively, by minimizing the Frobenius norm $\|V - WH\|^2_F$.
The rows of matrix $H$ represent the $K$ class vectors of the clustering, and $W_{m,k}$ contains the intensity of class $k$ on the $m$-th feature vector.
More importantly, it represents the intensity of class $k$ of that spatial location of a single tile.

On inference of a fresh matrix of feature vectors, the matrix $H$ stays fixed while the weight matrix $W$ is optimized in the same fashion.
To achieve clustering of the feature vectors, we assign to each feature vector the class with the highest intensity (weight).

\vspace{0em minus 0.5em}
\subsection{WSI-based clustering}

For the method to be applicable to microscopy images, we need to address several issues, primarily the tile overlap. 
The same patch in a slide is contained in multiple tiles (at different spatial locations), and therefore has multiple feature vectors associated with it. 
These feature vectors should be similar, as convolutional networks are mostly equivariant.  
However, this is not guaranteed, especially near the tile border. 
We address this problem by first omitting the 2px-thick outline of feature vectors from each tile and then computing the average of all feature vectors associated with each location $i,j$ (as noted, each feature vector comes from a different tile). 
When visualizing the clustering for the whole slide, these average feature vectors are used to determine in which cluster a given pixel lies. 
\looseness=-1

We also need to address the large size of WSIs, including the significant amount of background contained within these slides. 
We discard all tiles whose inner area of 256 × 256 pixels does not overlap with tissue.

\vspace{0em minus 0.5em}
\subsection{Evaluation Dataset and Model}

To evaluate our approach, we use the model for detecting prostate cancer from~\cite{gallo}.
This model is a binary classifier utilizing a VGG-16 backbone, followed by a global max-pool layer and a single fully connected layer.
It is trained to determine whether a patch's central area of 256 × 256 pixels  contains cancerous tissue.
Despite the model's relatively small size and simplicity, it achieves respectable performance on the test set, with 100\% slide-level accuracy~\cite{gallo}.  

This model was developed using the WSI dataset of hematoxylin- and eosin-stained patient prostate biopsies from the digital archive at the Department of Pathology, Masaryk Memorial Cancer Institute, Brno.
While each case is from a patient with a positive diagnosis, some of the biopsies are non-cancerous, yielding the positive/negative ratio of WSIs of 37/50.
All positive biopsies were manually annotated with polygons of cancerous areas.

The WSIs were tiled into overlapping patches of 512\texttimes 512 px with a stride of 256~px to reduce the risk of severing important morphological patterns, informed by expert knowledge of these patterns from a practicing pathologist.
We process the WSIs at level~1, which corresponds to 10\texttimes magnification and resolution 0.344~\textmu m/px.
The label for a patch is set as positive if the central area of the tile overlaps with the annotation.

\begin{figure}[ht]
    \centering
    \includegraphics[width=\linewidth]{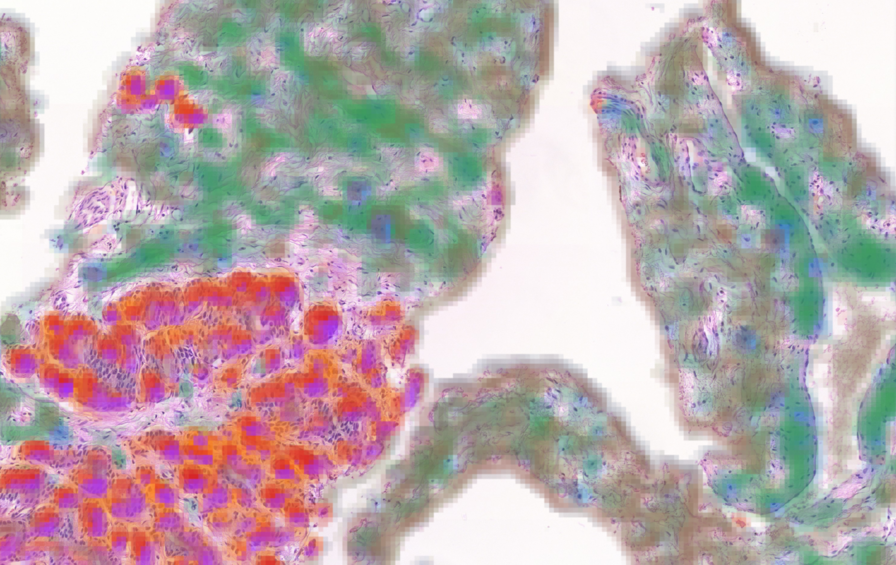}
    \caption{Visualisation produced by our technique, for 6 classes, overlapping intensities heatmaps. Classes are colored
    1\textcolor{cluster1}{$\blacksquare$},
    2\textcolor{cluster2}{$\blacksquare$},
    3\textcolor{cluster3}{$\blacksquare$},
    4\textcolor{cluster4}{$\blacksquare$},
    5\textcolor{cluster5}{$\blacksquare$}, and
    6\textcolor{cluster6}{$\blacksquare$}.}
    \label{fig:nmf6-all-soft}
    \vspace{-1.5em}
\end{figure}

To train the clustering models, we chose 16 positive and 8 negative slides from the dataset.
We selected the last (deepest) convolutional layer as the input layer to the clustering algorithm.
It produces outputs of dimensions 32\texttimes 32, with 512 channels.
As VGG-16 already uses ReLU, no modifications were necessary.

The training set consists of 37,332 tissue tiles (10,927 positive, 26,405 negative).
We trained three different models using $K=4$, $6$, and $8$.
Each trained clustering model was used for inference on the entire test dataset of 87 WSIs.
We generated heatmaps of intensities (weights) of each class, and a clustering overlay of the tissue.
These visualizations were presented to our resident pathologist to assess the quality of the models.

\vspace{-0em minus 1em}
\section{Results and Discussion}

\begin{figure}[ht]
    \centering
    \begin{tabular}{@{}c@{}c@{}}
        \includegraphics[width=0.5\columnwidth]{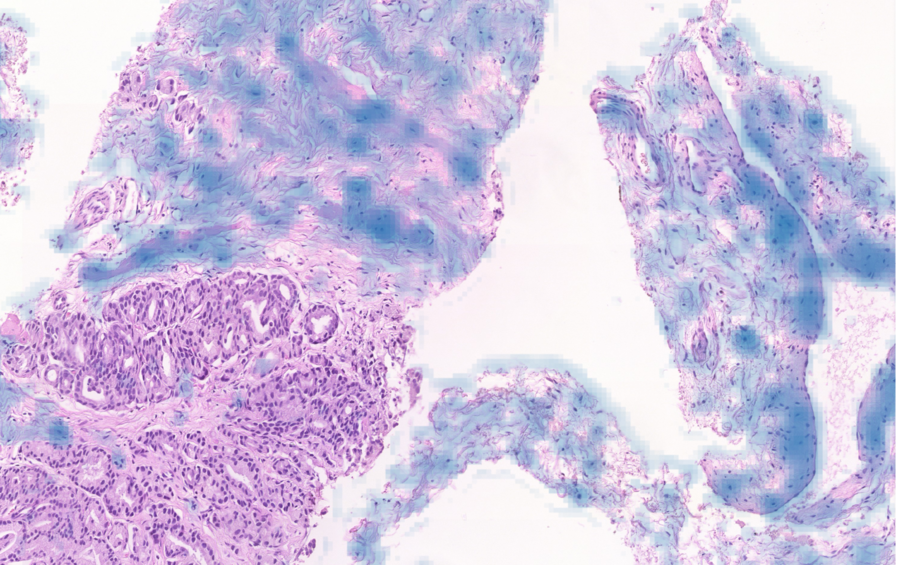} &
        \includegraphics[width=0.5\columnwidth]{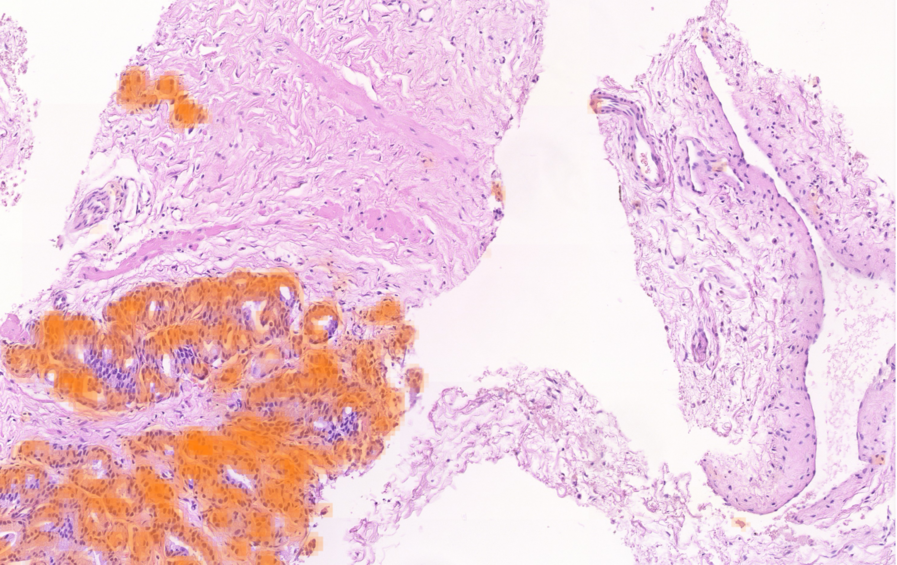} \\[-2pt]
        \includegraphics[width=0.5\columnwidth]{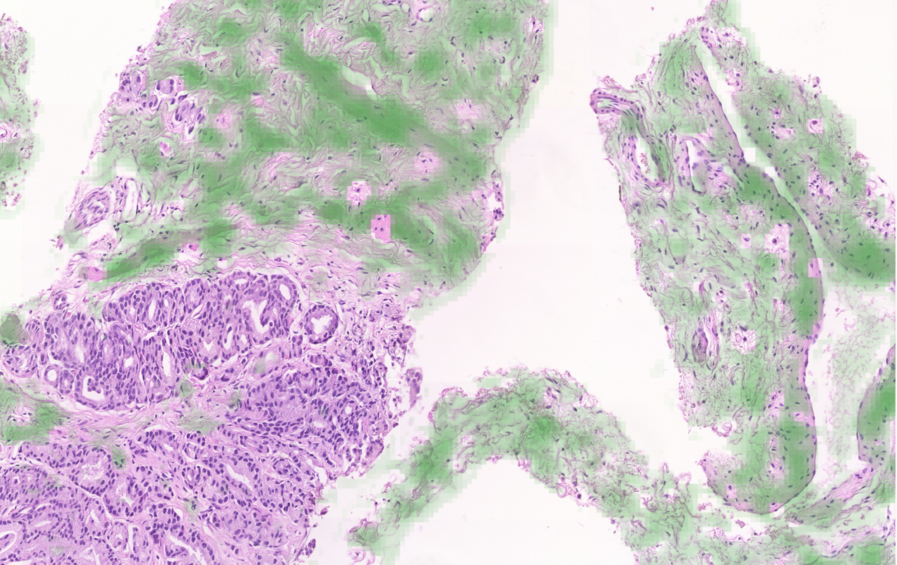} &
        \includegraphics[width=0.5\columnwidth]{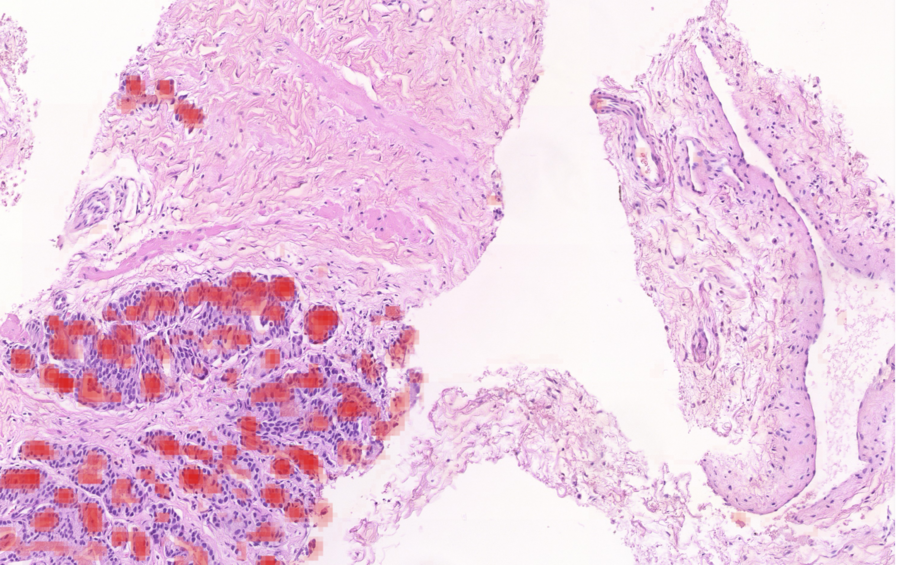} \\[-2pt]
        \includegraphics[width=0.5\columnwidth]{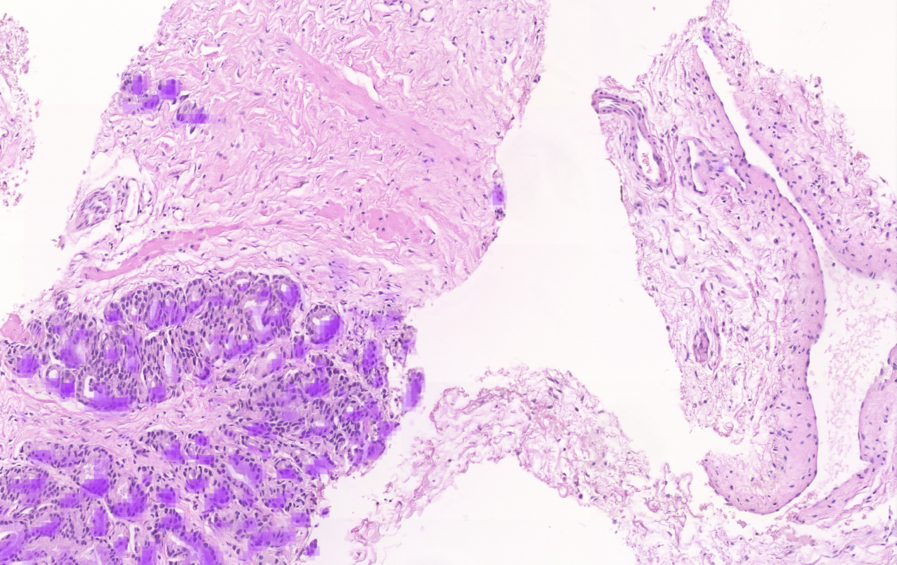} &
        \includegraphics[width=0.5\columnwidth]{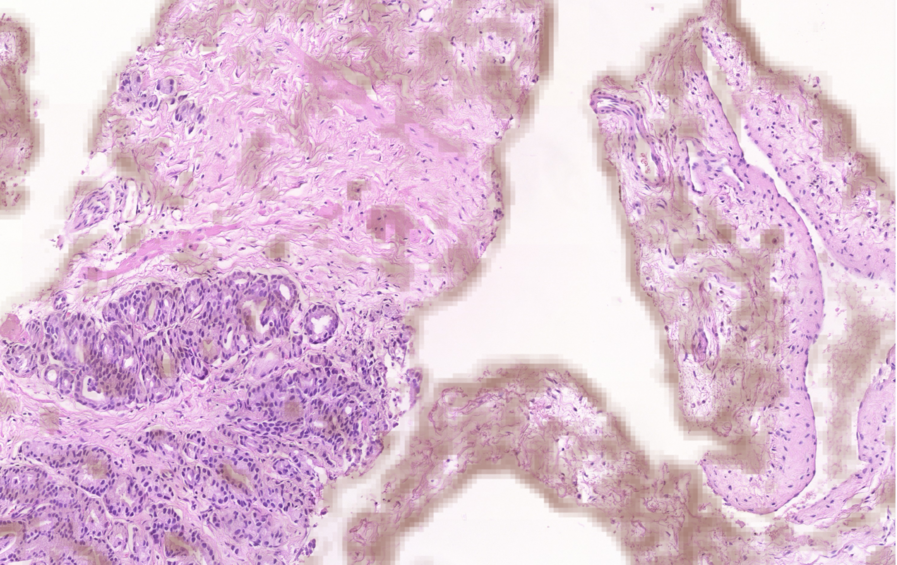}
    \end{tabular}
    \caption{Each of the 6 classes from Figure~\ref{fig:nmf6-all-soft}  visualised separately. Order is left-to-right. Classes are colored
    1\textcolor{cluster1}{$\blacksquare$},
    2\textcolor{cluster2}{$\blacksquare$},
    3\textcolor{cluster3}{$\blacksquare$},
    4\textcolor{cluster4}{$\blacksquare$},
    5\textcolor{cluster5}{$\blacksquare$}, and
    6\textcolor{cluster6}{$\blacksquare$}.
    }
    \label{fig:nmf6-separate-soft}
\end{figure}

Fig.~\ref{fig:nmf6-all-soft} shows a typical output of our model for $K=6$, where each of the classes is shown in a different color and classes are overlaid over the top of each other.
Fig.~\ref{fig:comp_mix} shows the intensity heatmap of all classes, and Fig.~\ref{fig:comp_hard} shows clustering on the same slide.
Fig.~\ref{fig:nmf6-separate-soft} shows the class intensities individually.
The pathologist observed that using higher values of~$K$ produced refined, coregistered clusters.
Using a lower $ K$ resulted in the merging of more correlated clusters, thereby decreasing granularity.

\vspace{-0em minus 1em}
\subsection{Pathologists evaluation}
\vspace{-0em minus 0.5em}

The resident pathologist manually inspected the heatmaps of the NMF model with $K = 6$ and provided the following description of the classes:
\begin{enumerate} \sl
    \item Fascicular structures with sparse elongated nuclei.
    \item Chains of densely packed nuclei; in tumour-bearing slides this pattern corresponds to carcinomatous areas, while in non-tumorous slides, it also captures normal epithelium.
    \item Similar to and overlapping with class~1, with emphasis on sparse tissue structure.
    \item Circular and/or semicircular small holes plus limited surroundings; this class is highly sensitive to very small lumina (carcinoma).
    \item Similar to and overlapping with class~2.
    \item Tissue edges plus limited surroundings.
\end{enumerate}
This suggests that the classes correspond to observable morphological structures.

\vspace{-1em}
\subsection{Quantitative analysis}
\vspace{-0em minus 0.5em}

We performed two tests to quantitatively verify these findings.
First measures the similarity of the classes, and the second measures the sensitivity of the classes to cancer.

\begin{figure}[!h]
    \centering
    \includegraphics[width=0.99\linewidth]{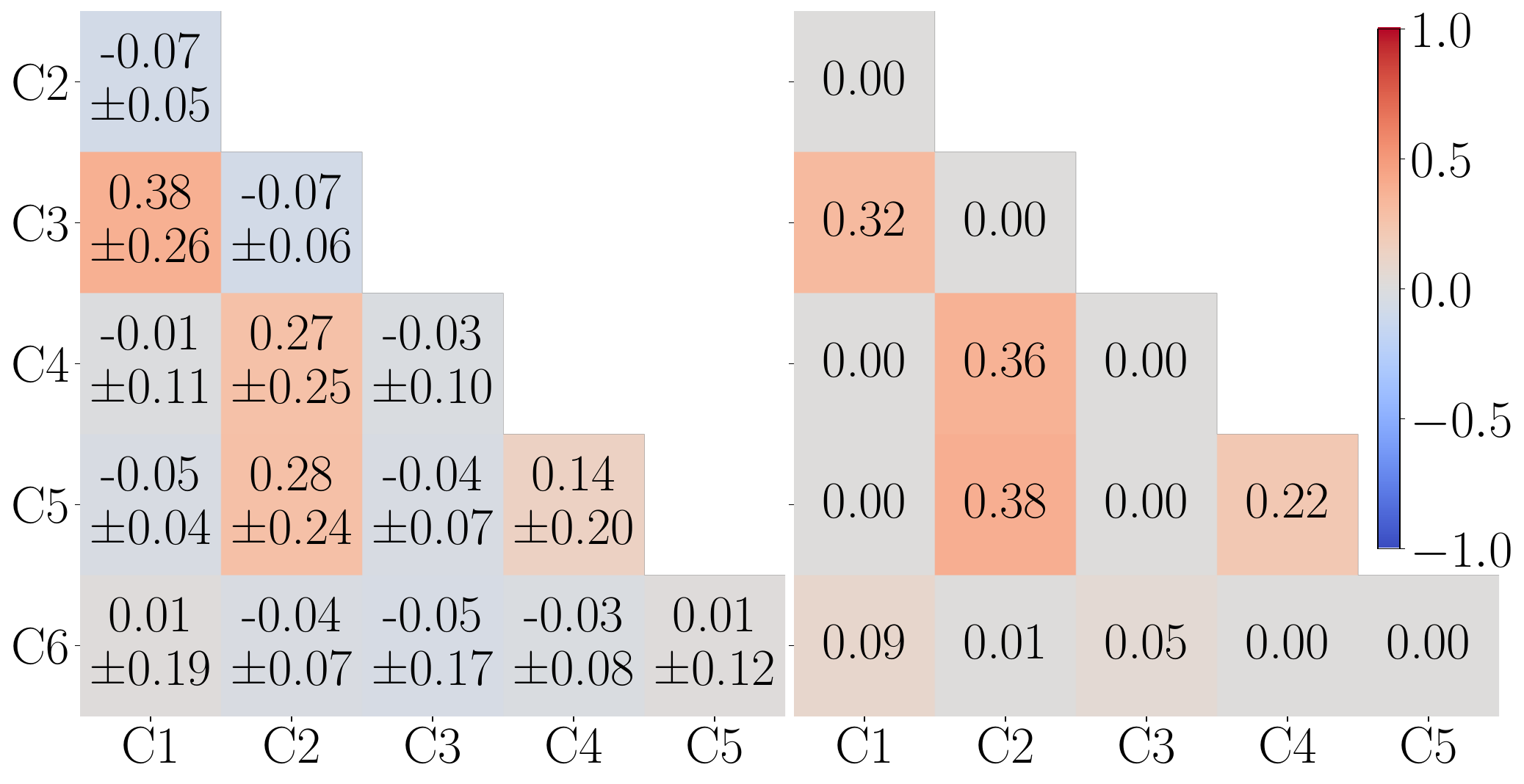}
    \caption{Correlation between class weights (mean $\pm$ std, left) and cosine similarity of class vectors (right).}
    \label{fig:weight_correlation}
    \vspace{-1em}
\end{figure}

For these tests, we used a random sample of 10,000 tiles (8,369 negative tiles, reflecting the class imbalance ) without replacement.
In~Fig.~\ref{fig:weight_correlation}, we show the mean correlation of the classes' weights on the tiles, together with the cosine similarity of the class vectors.
These results confirm the mutual closeness of classes 1, 3 and 2, 4, 5.
\begin{table}[b]
    \vspace{-1em}
    \small
    \centering
    \begin{tabular}[t]{cr}
        \toprule
        \textbf{Class} & \textbf{Coefficient}\\
        \midrule
        1 &  $-0.013$ \\
        2 &   $0.240$ \\
        3 &  $-0.068$ \\
        4 &   $0.462$ \\
        5 &   $0.368$ \\
        6 &  $-0.244$ \\
        \bottomrule
    \end{tabular}
        \qquad
    \begin{tabular}[t]{ll}
        \toprule
        \textbf{Metric} & \\
        \midrule
        Accuracy  & $0.984$ \\
        Precision & $0.970$ \\
        Recall    & $0.937$ \\
        F1 Score  & $0.954$ \\
        AUC Score & $0.998$ \\
        \bottomrule
    \end{tabular}    
    \caption{Class coefficients and statistics of a regression model predicting cancer positivity from the weights of classes.
    We confirm that classes 2, 4, and 5 highlight pro-cancer features.
    }
    \label{tab:regression}
    \vspace{-1em}
\end{table}

\begin{figure*}[!ht]
    \vspace{-1em}
    \begin{subfigure}[t]{0.32\linewidth}
      \includegraphics[width=\linewidth]{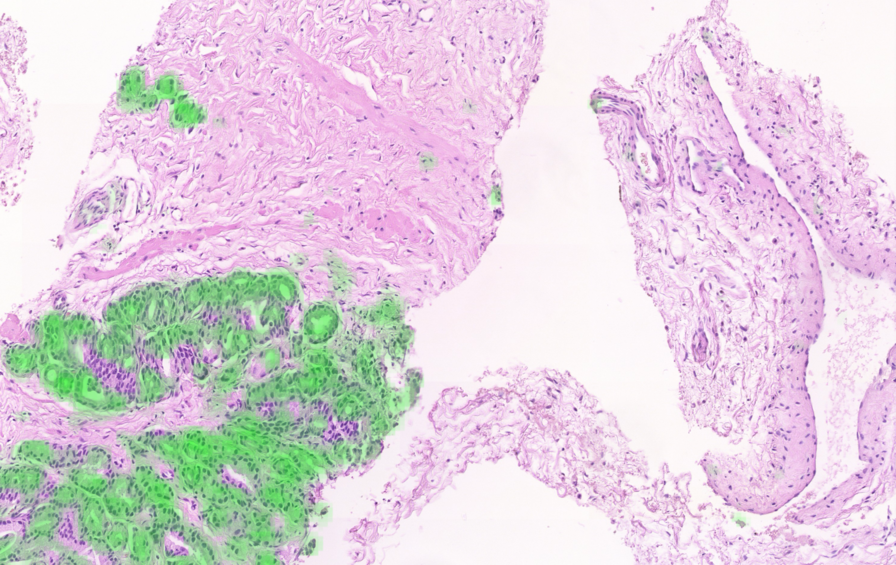}
        \caption{GradCAM}
        \label{fig:comp_grad}
    \end{subfigure}
    \begin{subfigure}[t]{0.32\linewidth}
      \includegraphics[width=\linewidth]{images/nmf6-mix.png}
        \caption{overlapping classes intensities}
        \label{fig:comp_mix}
    \end{subfigure}
    \begin{subfigure}[t]{0.32\linewidth}
      \includegraphics[width=\linewidth]{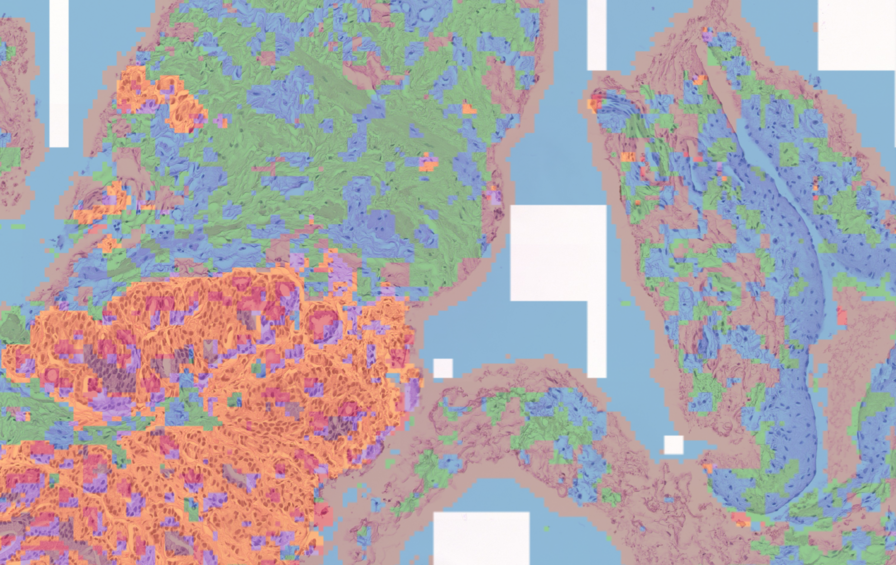}
        \caption{clustering}
        \label{fig:comp_hard}
    \end{subfigure}
    \caption{Visualising the same region using different methods. Classes at (b), (c) are colored
    1\textcolor{cluster1}{$\blacksquare$},
    2\textcolor{cluster2}{$\blacksquare$},
    3\textcolor{cluster3}{$\blacksquare$},
    4\textcolor{cluster4}{$\blacksquare$},
    5\textcolor{cluster5}{$\blacksquare$}, and
    6\textcolor{cluster6}{$\blacksquare$}.}
\end{figure*}

For the second test, we trained a logistic regression model that approximates the model's behavior based solely on clustering information.
We used the models' predictions (1 indicates cancer) as the ground truth and trained the linear model to predict cancer positivity from the sum of all weights of each class.
The statistics for the logistic regression are summarized in Tab.~\ref{tab:regression}.
We can see that the presence of classes 2, 4, and 5 strongly correlates with the tile being marked as positive, whereas 1, 3, and 6 tend to be more often present in negative tiles.
This is consistent with the descriptions of our resident pathologist.

We have experimented with other metrics for the linear model: using the coverage of the class, the maximal weight of each class, or the average positive weight, and found that the signs of the model weights are always identical.
We therefore report the one with the best overall scores.

\vspace{-0em minus 1em}
\subsection{Comparison with CAM-based methods}
\vspace{-0em minus 0.5em}

Our method is intended to provide global information about the model; in contrast, CAM-based methods highlight only relevant regions for prediction.

To compare these two, we measured intersection over union (IoU) of the regions with positive class weight and positive GradCAM weight.
We also measured the correlation of class weights with GradCAM weights.
The results are shown in Tab.~\ref{tab:GradCAM_IOU}.
We tried other CAM-based methods, such as GradCAM++ and HiResCAM, and the results are similar.

Classes 2, 4, and 5 strongly correlate with GradCAM explanation heatmaps, which is consistent with how they highlight morphological features indicative of cancer. 
However, the average overlap between GradCAM and these classes is low, indicating that our method provides more nuanced information.
A visual comparison is presented in Fig.~\ref{fig:comp_grad}, where green colour signifies a positive contribution to cancer positivity.

\section{Conclusion}

Our clustering-based explainability technique offers new insights into the inner workings of CNN-based digital pathology models. 
Unlike commonly used techniques such as GradCAM, the provided explanations focus on the global behavior of the model.
The explanations obtained by our method provide a means to increase confidence in CNN-based models, thereby facilitating their faster adoption in clinical practice.

The presented method is not limited to CNNs, but is applicable to any model architecture that internally works with spatial feature vectors.
Future work will focus on inspecting large foundational transformer models for digital pathology, as well as other models, while evaluating how well this methodology generalizes to other domains.

\begin{table}[h]
    \small
    \centering
    \begin{tabular}{
            c
            S[table-format=1.2(2)]
            S[table-format=1.2(2)]
            S[table-format=1.2(2)]
        }
        \toprule
        {\textbf{Class}} & {\textbf{IoU}} & {\textbf{Corr Benign}} & {\textbf{Corr Cancer}} \\
        \midrule
        1 & 0.00 \pm 0.01 & -0.07 \pm 0.04 & -0.13 \pm 0.05 \\
        2 & 0.27 \pm 0.12 &  0.65 \pm 0.27 &  0.85 \pm 0.11 \\
        3 & 0.01 \pm 0.02 & -0.07 \pm 0.04 & -0.14 \pm 0.07 \\
        4 & 0.11 \pm 0.10 &  0.39 \pm 0.34 &  0.64 \pm 0.14 \\
        5 & 0.15 \pm 0.10 &  0.50 \pm 0.30 &  0.56 \pm 0.12 \\
        6 & 0.01 \pm 0.01 & -0.05 \pm 0.04 & -0.10 \pm 0.07 \\
        \bottomrule
    \end{tabular}
    \caption{Mean and std of IoU between clusters and GradCAM mask.
    The second and third columns show the correlation between class weights and the GradCAM heatmap, split by the predicted label.}
    \label{tab:GradCAM_IOU}
    \vspace{-1em}
\end{table}

\section{Compliance with ethical standards}
\label{sec:ethics}

The project was approved by the Ethical Committee of Masaryk Memorial Cancer Institute, No. MOU 385 920.

\section{Acknowledgments}
\label{sec:acknowledgments}

The work utilized the infrastructure and AI framework developed BioMedAI, supported by the European Union’s Horizon Europe research and innovation program under No.~101079183 (BioMedAI TWINNING).
Computational resources were provided by the e-INFRA CZ project (ID:90140), supported by the Ministry of Education, Youth, and Sports of the Czech Republic.
The work on this paper was supported by the Czech Science Foundation (GAČR) grant no. 26-23981S

The authors have no relevant financial or non-financial interests to disclose.


\end{document}